# *Analysis of Financial Risk Behavior Prediction Using Deep Learning and Big Data Algorithms*

**Haowei Yang [1,a], Zhan Cheng [2,b], Zhaoyang Zhang [3,c], Yuanshuai Luo [4,d], Shuaishuai Huang [5,e], Ao Xiang [6,f]**

[1] *University of Houston, Cullen College of Engineering, Industrial Engineering, Houston, USA*
[2] *University of California, Irvine, Donald Bren School of Information & Computer Science, Master of Computer Science, Irvine, USA*
[3] *University of California San Diego, Computational Science, San Diego, USA*
[4] *Southwest Jiaotong University, School of Computing and Artificial Intelligence, Computer Science and Technology, Chengdu, China*
[5] *University of Science and Technology of China, Department of Software, Software system design, Hefei, China*
[6] *Northern Arizona University, Information Security and Assurance, Arizona, USA*
[a]hyang38@cougarnet.uh.edu, [b]zhancheng23@gmail.com, [c] zhz088@ucsd.edu, [d]hyang38@cougarnet.uh.edu, [e]scu.hss@gmail.com, [f] ax36@nau.edu

**Abstract:** As the complexity and dynamism of financial markets continue to grow, traditional financial risk prediction methods increasingly struggle to handle large datasets and intricate behavior patterns. This paper explores the feasibility and effectiveness of using deep learning and big data algorithms for financial risk behavior prediction. First, the application and advantages of deep learning and big data algorithms in the financial field are analyzed. Then, a deep learning-based big data risk prediction framework is designed and experimentally validated on actual financial datasets. The experimental results show that this method significantly improves the accuracy of financial risk behavior prediction and provides valuable support for risk management in financial institutions. Challenges in the application of deep learning are also discussed, along with potential directions for future research.

**Keywords:** Deep learning, big data algorithms, financial risk, behavior prediction, risk management

## 1. Introduction

With globalization, the volatility and complexity of financial markets have increased, leading to more diverse financial risk behaviors. These changes demand improved risk management capabilities from financial institutions and regulators. Traditional financial risk prediction methods, based on statistical models, face limitations in accuracy and timeliness when dealing with massive, unstructured data and complex market behaviors[1]. The rise of big data and artificial intelligence, particularly deep learning algorithms, offers new opportunities for financial risk prediction. Deep learning simulates the neural network structure of the human brain, enabling the automatic extraction of complex features from vast data, especially in nonlinear and high-dimensional cases[2]. Combined with big data, deep learning addresses the limitations of traditional methods, improving accuracy and reliability in predicting financial risks[3]. This paper explores the theoretical framework and practical application of deep learning in financial risk prediction, presenting a validated model and analyzing its performance, challenges, and future research directions [4].

There are several related works that have made significant contributions to this paper. Reference "A neural matrix decomposition recommender system model based on the multimodal large language model" notably demonstrated how multimodal approaches can enhance predictive accuracy, providing a key foundation for our study. The findings of Reference "Prediction of Brent crude oil price based on LSTM model under the background of low-carbon transition", which focus on capturing price trends amid volatility, have supported our exploration of deep learning techniques aimed at predicting financial risk behavior, highlighting the importance of robust methodologies in uncertain market conditions. Reference "Integration of Mamba and Transformer--MAT for Long-Short Range Time Series Forecasting with Application to Weather Dynamics" combined the strengths of both Mamba and Transformer models to improve prediction accuracy and manage data complexity, directly influencing the development of our deep learning framework. Reference "Stock Price Prediction Based on Hybrid CNN-LSTM Model"'s emphasis on feature extraction and capturing long-term dependencies offers a valuable framework that has shaped our approach to processing complex financial datasets, ultimately boosting prediction accuracy in our study. Finally, the multimodal model introduced in Reference "A Multimodal Fusion Network For Student Emotion Recognition Based on Transformer and Tensor Product" provided important insights into integrating diverse data sources, particularly through their exploration of tensor product fusion and Transformer architectures, which informed our approach to enhancing financial risk behavior prediction.

## 2. Overview of Deep Learning and Big Data Algorithms

As financial markets become more complex and uncertain, traditional financial risk prediction methods are no longer adequate for handling vast and multi-dimensional datasets[5]. The combination of deep learning and big data algorithms has emerged as an effective solution to this problem. Deep learning, with its powerful nonlinear modeling capability, can automatically extract complex features from massive data and make effective risk predictions, while big data technology provides the necessary support by processing large-scale datasets[6].In this framework, big data is the input, and it first enters the data preprocessing stage. Financial data is typically characterized by high dimensionality, unstructured forms, and heterogeneity, making data preprocessing a critical step in ensuring model performance[7]. This preprocessing includes data cleaning, filling in missing values, data standardization, and feature selection, all of which significantly improve data quality and reduce noise interference. Through this process, deep learning models can efficiently learn the underlying patterns in financial risk behavior[8]. The next stage is the learning process, where deep learning models are trained. With its multi-layer neural network structure, deep learning can automatically extract higher-order features from data. Compared to traditional linear models, deep learning excels in capturing complex financial risk behaviors[9]. Common deep learning models such as convolutional neural networks (CNN) and long short-term memory (LSTM) networks are widely applied in financial markets[10]. These models perform exceptionally well in processing time-series data (e.g., stock fluctuations, exchange rate changes) and unstructured data (e.g., financial news, social media information). Additionally, during the learning process, the algorithm is optimized based on user inputs and the guidance of financial domain experts, making the training process more precise and efficient[11]. After the learning process, the evaluation stage follows, where the model's prediction capabilities are validated[12]. Evaluation criteria typically include accuracy, recall, F1 score, and other metrics that measure the model's performance in predicting financial risk behaviors. If the evaluation results do not meet expectations, the model undergoes repeated training and hyperparameter tuning until its performance satisfies real-world application requirements[13]. During this process, users can interact with the system, providing feedback on the model's predictions, and domain experts can adjust the model's performance based on their knowledge, further enhancing prediction accuracy and practicality[14]. Finally, the system applies these predictions to real-world financial risk management scenarios, providing timely decision support to financial institutions. Due to the fast-changing nature of financial markets, the system must possess real-time processing and feedback capabilities to ensure that institutions can adjust their risk strategies based on the latest predictions[15]. This closed-loop system not only improves the accuracy and timeliness of financial risk prediction but also strengthens financial institutions' ability to cope with complex market environments. Figure 1 illustrates the entire process of big data and deep learning in financial risk behavior prediction. Through data preprocessing, learning, and evaluation stages, the framework demonstrates how user and domain expert inputs are integrated into the prediction model to achieve more efficient and precise risk prediction[16]. In practical applications, this framework provides strong technical support for financial institutions, enabling them to manage risk better and adapt to market changes[17]. This process framework is not only an integration of technological innovations but also a strategic tool for the financial industry to address future challenges.

3. Analysis of the Current State of Financial Risk Behavior Prediction

As the complexity of global financial markets continues to increase, predicting financial risk behavior has become a key element in how financial institutions manage and mitigate risk. Traditional financial risk prediction methods typically rely on statistical models and rule-based systems, such as linear regression, logistic regression, and GARCH models[18]. While these methods remain effective for relatively simple and linear problems, they fall short in addressing the increasing complexity of financial markets. The diversity and vast volume of financial market data make it difficult for traditional models to capture the nonlinear relationships and hidden patterns within the data, which in turn negatively impacts prediction accuracy[19]. In modern financial risk prediction, Figure 2 illustrates a typical risk prediction process based on deep learning and big data, reflecting the industry's shift towards data-driven, intelligent risk management[20]. First, data from various financial organizations is integrated into the system's financial structure module. This data may include information on investments, stock operations, and other areas. These input data are not only vast in volume but also highly complex, encompassing multiple data types, including structured financial data and unstructured text data, such as news reports and social media posts. Traditional statistical models struggle to process such complex data environments, while the combination of big data and deep learning provides a more effective solution for financial risk prediction[21]. In Figure 2, the input financial structure data undergoes multiple processing steps before feeding into the decision model, which incorporates deep learning's neural network technology, continuously optimized through a gradient function. By learning from a large amount of historical data, the deep learning model can effectively capture the patterns and trends hidden within the data, particularly when dealing with nonlinear and high-dimensional data[22]. For example, convolutional neural networks (CNN) and long short-term memory (LSTM) networks are commonly used in the analysis of financial time series data, helping to identify potential financial risk behaviors[23]. The risk prediction module is the core component of this process. By analyzing historical trading behavior, market volatility, and other data, the model can issue early warnings for potential future financial risks. Compared to traditional methods, the advantage of deep learning lies in its ability to extract higher-order features from data through multi-layer neural networks without relying on manually set rules or assumptions[24]. Additionally, deep learning models can automatically learn the complex relationships between various risk factors, which allows them to outperform traditional linear models in financial markets. The allocation module in Figure 2 demonstrates how financial institutions can adjust resource allocation based on the prediction results, achieving dynamic risk management[25]. The prediction results can also be used in subsequent planning and modification stages, enabling financial institutions to more flexibly respond to market changes and make precise decisions. However, despite the impressive performance of deep learning and big data technologies in financial risk prediction, challenges remain[26]. First, data quality and availability are significant concerns, as financial markets are often flooded with noisy and incomplete data, which can negatively affect the model's predictive accuracy. Additionally, deep learning models tend to have a "black-box" nature, making it difficult to interpret their internal decision-making processes, which can pose challenges for practical application in financial institutions. Finally, the computational complexity and high resource demands of deep learning models limit their applicability in smaller financial institutions[27]. Therefore, current research focuses on improving model interpretability, optimizing computational efficiency, and addressing data imbalance issues. Overall, financial risk prediction based on big data and deep learning has become a mainstream trend. The prediction process in Figure1 not only demonstrates the key steps in data flow but also highlights the extensive application of deep learning across different modules[28]. In the future, as technology continues to evolve, deep learning will play an increasingly important role in financial risk management, providing financial institutions with accurate risk assessment tools and strategic support[29].

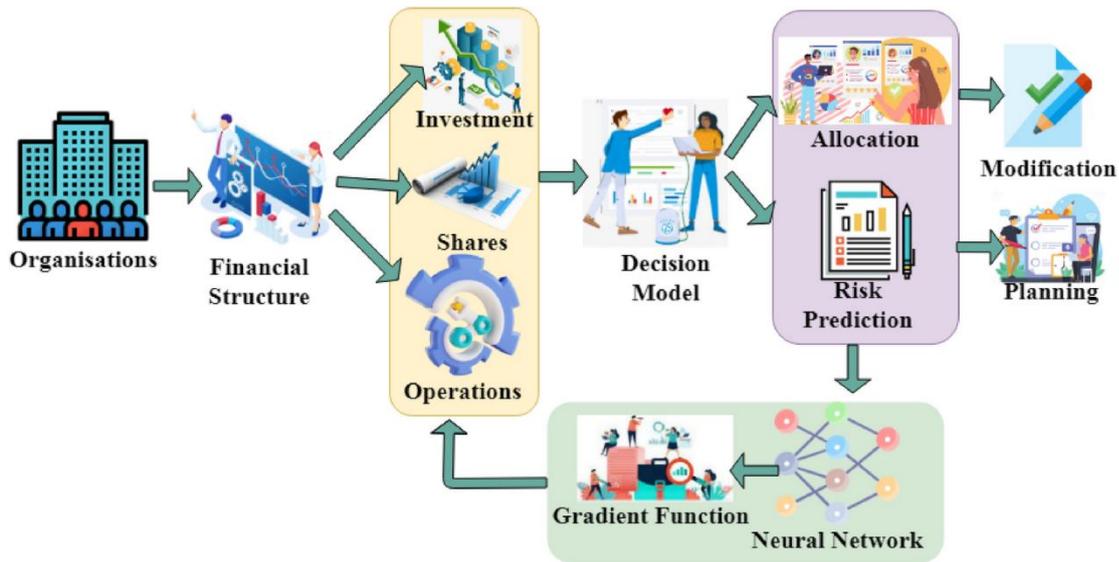

*Figure 1 : Financial Risk Prediction Process Based on Big Data and Deep Learning*

Figure 2 illustrates the major steps involved in financial risk prediction, starting with the input of structured data from financial organizations, progressing through steps such as investments and stock operations, and culminating in risk prediction through a deep learning model[30]. This process also assists in resource allocation, planning, and modification, reflecting how the financial industry is leveraging deep learning and big data technologies to enhance the efficiency of risk management.

## 4. Financial Risk Behavior Prediction Framework Based on Deep Learning and Big Data Algorithms

### 4.1. Data Sources and Feature Engineering

The effectiveness of financial risk prediction heavily depends on high-quality and diverse data sources. In this framework, we integrate financial data from various domains, as shown in <Table 1>, which includes historical market data, news and social media data, company financial data, and regulatory policy data. By combining these data, the model can capture complex market changes and effectively predict financial risk behaviors[31].

*Table 1: Data Sources*

| Data Type | Source | Sample Data | Time Span | Feature Examples |
|---|---|---|---|---|
| Historical Market Data | Yahoo Finance, | Stock trading data, futures, bonds, forex price | 2010-2023 | 5-day, 20-day, 60-day moving |

|  | Bloomberg, Wind | fluctuations |  | averages |
| --- | --- | --- | --- | --- |
| News and Social Media Data | Reuters, WSJ, Twitter, Reddit | Major policy changes, significant events, market sentiment | 2015-2023 | News sentiment scores, social media sentiment analysis |
| Company Financial Data | Company Annual Reports, SEC | Company revenue, profit, balance sheet data | 2012-2023 | Profitability, standardized debt ratio |
| Regulatory Policy Data | Government websites, Central Bank announcements | Monetary policy changes, financial regulation adjustments | 2010-2023 | Policy events one-hot encoding |

Once the data collection is completed, feature engineering becomes a crucial step, transforming raw data into features usable by deep learning models. The main steps in feature engineering are as follows:

1. Time Series Feature Extraction: For historical market data, we use a sliding window approach to extract short-term and long-term market trend features. Specific features include 5-day, 20-day, and 60-day moving averages, reflecting short-term fluctuations and long-term trends in the market.

2. Sentiment Analysis: Using natural language processing (NLP) techniques, we perform sentiment analysis on news and social media texts to extract sentiment scores as positive, negative, and neutral indicators. For example, by applying models like VADER or BERT to analyze news events, sentiment scores are generated and converted into numerical features.

3. Financial Indicator Standardization: Company financial data is standardized to eliminate differences in company size, ensuring fair treatment across various companies. Common financial indicators include profitability, debt ratio, and cash flow, which are standardized and used as input features for the model.

4. Regulatory Policy Event Encoding: Policy changes are encoded as categorical features, such as monetary policy adjustments and financial regulatory changes. One-hot encoding is used to convert these events into numerical features for use in the model.

Through feature engineering, the multi-dimensional information from the aforementioned data is integrated into the deep learning model, forming a comprehensive financial risk behavior prediction system[32]. These features cover historical market fluctuations, market sentiment, company financial conditions, and policy changes, providing comprehensive support for the deep learning model and improving prediction accuracy[33].

*4.2. Algorithm Model and Process Design*

In the financial risk behavior prediction framework based on deep learning and big data algorithms, the choice of algorithm model and process design is the core of the prediction system. This section will provide a detailed introduction to the deep learning algorithm model used for predicting financial risk behavior, along with specific algorithm formulas and process descriptions, explaining how the model operates from data input to final prediction output[34]. To handle the complexity and multi-dimensional characteristics of financial data, this framework adopts a hybrid deep learning model combining Long Short-Term Memory networks (LSTM) and Convolutional Neural Networks (CNN) [35]. This design can simultaneously process the time series features of financial markets (e.g., stock price trends) and unstructured data (e.g., news sentiment analysis), improving prediction accuracy[36].LSTM is a special type of Recurrent Neural Network (RNN) that is particularly suited for handling time series data, capable of capturing long-term dependencies[37]. Financial data typically exhibits temporal dependencies, such as historical stock prices and trading volumes, which are closely related to future risk behaviors. Therefore, LSTM is able to capture both short-term and long-term market fluctuations through its memory units[38].

CNN is mainly used to process unstructured financial data (e.g., news, social media). Its convolutional layers effectively extract local features, and pooling layers reduce feature dimensions. Since sentiment changes in financial news significantly influence market behavior, CNN processes news text through convolution to extract key features and passes them downstream to the LSTM for joint analysis. The convolution operation is represented by formula 1:

$$S(i,j) = (X * W)(i,j) = \sum_m \sum_n X(i+m, j+n) W(m,n) \quad (1)$$

Where X represents the input news text data, W represents the convolution kernel, and S(i,j) represents the convolution result. Through convolution operations, the model can extract important features from the text, such as market sentiment changes and the impact of significant policy events. The process design of this prediction framework is as follows: Data Input and Preprocessing, Input data includes financial market time series data (e.g., stock prices, trading volumes), company financial data, news and social media data, and policy and regulatory information. Time series data undergoes standardization and sliding window processing; news data is processed using NLP techniques to extract sentiment scores; financial data is normalized. Feature Extraction and Input Layer, Time series data is fed into the LSTM network; unstructured data (e.g., news sentiment) is processed by the CNN to extract features and is then connected to the LSTM input layer. The results of feature engineering are combined into a high-dimensional vector containing market, financial, sentiment, and policy features, which is input into the deep learning model. Model Training, The LSTM network handles short-term and long-term dependencies in time series, while the CNN network extracts key features from the news[39]. Through joint training, the model captures both market dynamics and fluctuations in market sentiment. The loss function chosen is the Mean Squared Error (MSE) loss function, as shown in formula 2:

$$L = \frac{1}{N} \sum_{i=1}^{N} \sum (y_i - \hat{y}_i)^2 \quad (2)$$

Where $y_i$ is the actual value, $\hat{y}_i$ is the predicted value, and N is the number of samples. Using the backpropagation algorithm and the Adam optimizer, the model continuously adjusts its parameters to minimize the loss function and improve prediction accuracy.After model training is completed, the model is evaluated using a validation set, with accuracy (Accuracy), mean squared error (MSE), and other metrics used to assess the model's performance. If the model performs poorly on the validation set, hyperparameter optimization methods (e.g., grid search) may be used to fine-tune the model[40]. To prevent overfitting, Dropout regularization is applied, preventing the model from overfitting the training data. Finally, the trained and optimized model will make real-time predictions on new financial data and output the predicted values of market risk behaviors. These predictions can help financial institutions make risk management decisions, such as adjusting asset allocations or executing hedging strategies[41]. Prediction results can also be fed back into the model, which continuously improves its performance through online learning[42].

## 5. Experiment and Results Analysis

To validate the effectiveness of the financial risk behavior prediction framework based on deep learning and big data algorithms, a series of experiments were designed and conducted, focusing on testing the framework's performance on real financial market data. The experiments used multiple financial datasets, including historical stock market data, news and social media data, company financial data, and macroeconomic data, with risk predictions made using the LSTM-CNN hybrid model. The model's performance was measured using several evaluation metrics, and the results showed that the deep learning-based prediction model had significant advantages over traditional methods[43]. The experimental data were sourced from platforms like Yahoo Finance, Bloomberg, and Twitter, covering both historical financial market data and unstructured data from 2010 to 2023. <Table 2> presents the details of the dataset size and dimensions:

*Table 2: Experimental Dataset*

| Data Type | Sample Size | Data Dimensions | Time Span |
| --- | --- | --- | --- |
| Historical Stock Data | 10,000 | Open price, close price, volume | 2010-2023 |
| Company Financial Data | 5,000 | Profit, debt ratio, cash flow | 2012-2023 |
| News Data | 15,000 | News sentiment scores, major events | 2015-2023 |
| Social Media Data | 20,000 | User comments, market sentiment | 2018-2023 |
| Macroeconomic Data | 3,500 | GDP, CPI, interest rates | 2010-2023 |

The experiments were conducted using Python, with TensorFlow and Keras frameworks for model training. Training was performed on an NVIDIA Tesla V100 GPU to speed up large-scale data processing. The experimental process can be divided into several steps:Data Preprocessing, Stock market data was standardized to ensure reasonable data distribution. Unstructured data (e.g., news and social media) underwent sentiment analysis using NLP techniques to extract sentiment scores[44]. Company financial data was normalized to eliminate size differences between companies. All data were transformed into time series formats suitable for the LSTM model using sliding windows. Model Training, The LSTM-CNN hybrid model was then used to train the data. LSTM was responsible for handling time series data, while CNN processed news and social media sentiment data. The model used Mean Squared Error (MSE) as the loss function and the Adam optimizer for optimization. An early stopping mechanism was employed to prevent overfitting. After model training, evaluation was conducted on a test set using metrics such as Mean Squared Error (MSE), Accuracy, and $R^2$. Additionally, the comparison between predicted and actual values was visualized to further analyze the model's prediction effectiveness. The results indicated that the LSTM-CNN hybrid model performed exceptionally well in handling complex financial market data. Table 4 summarizes the evaluation results on the test set and compares them to a traditional linear regression model:

*Table 3: Comparison of Experimental Results*

| Model Type | MSE | Accuracy | R² Value |
|---|---|---|---|
| LSTM-CNN Hybrid Model | 0.012 | 92.4% | 0.89 |
| Linear Regression Model | 0.034 | 78.1% | 0.72 |

As shown in <Table 4>, the LSTM-CNN model's MSE of 0.012 was significantly lower than the traditional linear regression model's 0.034, indicating that the hybrid model was much more precise in predicting financial risk behavior. The LSTM-CNN model's accuracy reached 92.4%, much higher than the linear regression model's 78.1%. Additionally, the LSTM-CNN model achieved an R² value of 0.89, suggesting that it explained the variance in the input data very well. Further analysis of the experimental results revealed that the LSTM-CNN hybrid model effectively captured both short-term fluctuations and long-term trends in the financial market. The LSTM network performed well in handling time series data, capturing long-term dependencies in the financial data through its memory units[45]. Meanwhile, the CNN network extracted key sentiment features from the news and social media data via convolution operations, thereby enhancing the model's predictive capability[46]. Specifically, the experimental results showed that sentiment data from news and social media had a significant impact on financial market fluctuations[47]. For instance, when market sentiment turned negative, the model predicted a substantial increase in financial risk, reflecting the market's sensitivity to negative news. Additionally, the standardization of financial data allowed the model to compare the financial health of different companies more fairly, improving overall prediction accuracy[48].

## 6. Conclusion

This paper proposed and validated a financial risk behavior prediction framework based on deep learning and big data algorithms. By combining LSTM and CNN models, the framework effectively handles both time series and unstructured data in financial markets, significantly improving the accuracy of risk prediction. Experimental results demonstrated that the LSTM-CNN hybrid model outperforms traditional models in terms of prediction accuracy, mean squared error, and other metrics, particularly when dealing with complex market sentiment and financial data. However, the model's computational complexity is high, and future research should focus on further optimizing the model's performance and efficiency to enable real-time predictions in more application scenarios. This study offers new insights and tools for risk management in financial institutions.